%% file: miccai23.tex
\documentclass[runningheads]{llncs}
\usepackage{cite}
\usepackage{amsmath,amssymb,amsfonts}
\usepackage{algorithmic}
\usepackage{graphicx}
\usepackage{textcomp}
\usepackage{xcolor}
\usepackage{multirow}
\usepackage{hyperref}
\begin{document}
\title{A Novel Confidence Induced Class Activation Mapping for MRI Brain Tumor Segmentation}
%
%
\author{
Yu-Jen Chen\inst{1} \and
Yiyu Shi\inst{2} \and
Tsung-Yi Ho\inst{3}}
%
\institute{National Tsing Hua University, Hsinchu, Taiwan \and
University of Notre Dame, IN, USA \and
The Chinese University of Hong Kong, Hong Kong
\\
\email{yujenchen@gapp.nthu.edu.tw}}
\maketitle              
\begin{abstract}

\input{Section/Abstract.tex}
\end{abstract}

\section{Introduction}
\label{sec:intro}
\input{Section/Introduction.tex}


\section{Confidence-induced CAM (Cfd-CAM)}
\label{sec:methodology}

\input{Section/Methodology.tex}

\section{Experiments}
\label{sec:experiments}
\input{Section/Experiments.tex}

\section{Results}
\label{sec:results}
\input{Section/Results.tex}

\section{Conclusion}
\label{sec:conclusion}
\input{Section/Conclusion.tex}
%
%
%
\bibliographystyle{splncs04}
\bibliography{miccai23,vincent_paper}

\end{document}


\input{Table/Supple_Res50}

\input{Table/Supple_cfd}

%% file: Section/Abstract.tex
Magnetic resonance imaging (MRI) is a commonly used technique for brain tumor segmentation, which is critical for evaluating patients and planning treatment. To make the labeling process less laborious and dependent on expertise, weakly-supervised semantic segmentation (WSSS) methods using class activation mapping (CAM) have been proposed.
However, current CAM-based WSSS methods generate the object localization map using internal neural network information, such as gradient or trainable parameters, which can lead to suboptimal solutions.
To address these issues, we propose the confidence-induced CAM (Cfd-CAM), which calculates the weight of each feature map by using the confidence of the target class.
Our experiments on two brain tumor datasets show that Cfd-CAM outperforms existing state-of-the-art methods under the same level of supervision.
Overall, our proposed Cfd-CAM approach improves the accuracy of brain tumor segmentation and may provide valuable insights for developing better WSSS methods for other medical imaging tasks.

%% file: Section/Introduction.tex
Deep learning \cite{chen2020zero,wen2020noises,chen2021ct,chen2021one,chen2022representative} has shown great potential in various of application for medical images. Among all of them, medical image segmentation is a critical task in disease diagnosis and assessment. Most existing approaches automating this process by extracting the location information of specific tissues or substances. However, most deep-learning-based approaches require fully or partially labeled training datasets, which can be time-consuming and expensive to annotate.

To address this issue, recent efforts have focused on developing segmentation frameworks with little or no segmentation labels. Weakly-Supervised Semantic Segmentation (WSSS) has emerged as a promising approach that utilizes weak supervision, such as image-level classification labels. The majority of works in this direction have centered around Class Activation Mapping (CAM) \cite{zhou2016learning}, which was originally developed as a technique for neural network visualization.

However, existing CAM-based WSSS methods \cite{zhou2016learning,selvaraju2017grad,omeiza2019smooth,wang2020score,ramaswamy2020ablation,wang2020self,lee2021lfi,zhang2018adversarial,tang2021m} generate the activation map from internal neural network information, which can lead to suboptimal solutions. For example, Grad-CAM\cite{selvaraju2017grad} calculates the gradients between the target class confidence and the feature map and uses them as weights. Wang et al. \cite{wang2020score} proposed the ScoreCAM, which showed that feature maps with higher weights may still have lower contributions to the prediction score.

In this paper, we propose a confidence-induced CAM (Cfd-CAM) for brain tumor segmentation in magnetic resonance imaging (MRI). Unlike recent CAM methods, Cfd-CAM uses the classification confidence of the input image being masked by its feature maps to estimate the importance of each feature map properly. Our code is publicly available at \\ \href{https://github.com/windstormer/Cfd-CAM}{https://github.com/windstormer/Cfd-CAM}

Experimental results on the 2021 Brain Tumor Segmentation Challenge (BraTS 2021) \cite{menze2014multimodal,bakas2017advancing,bakas2018identifying} and the TCGA-LGG dataset \cite{pedano2016radiology} demonstrate the superiority of Cfd-CAM over existing CAM methods in extracting segmentation results from classification networks. To ensure a fair comparison, we also reproduce a state-of-the-art unsupervised segmentation method and two supervised approaches as the performance ceiling. Our proposed Cfd-CAM approach shows great promise in reducing the reliance on fully or partially labeled training datasets for medical image segmentation, and we believe it may have broader applications in other fields.

%% file: Section/Methodology.tex
\begin{figure}
\centering
\includegraphics[width=0.95\linewidth]{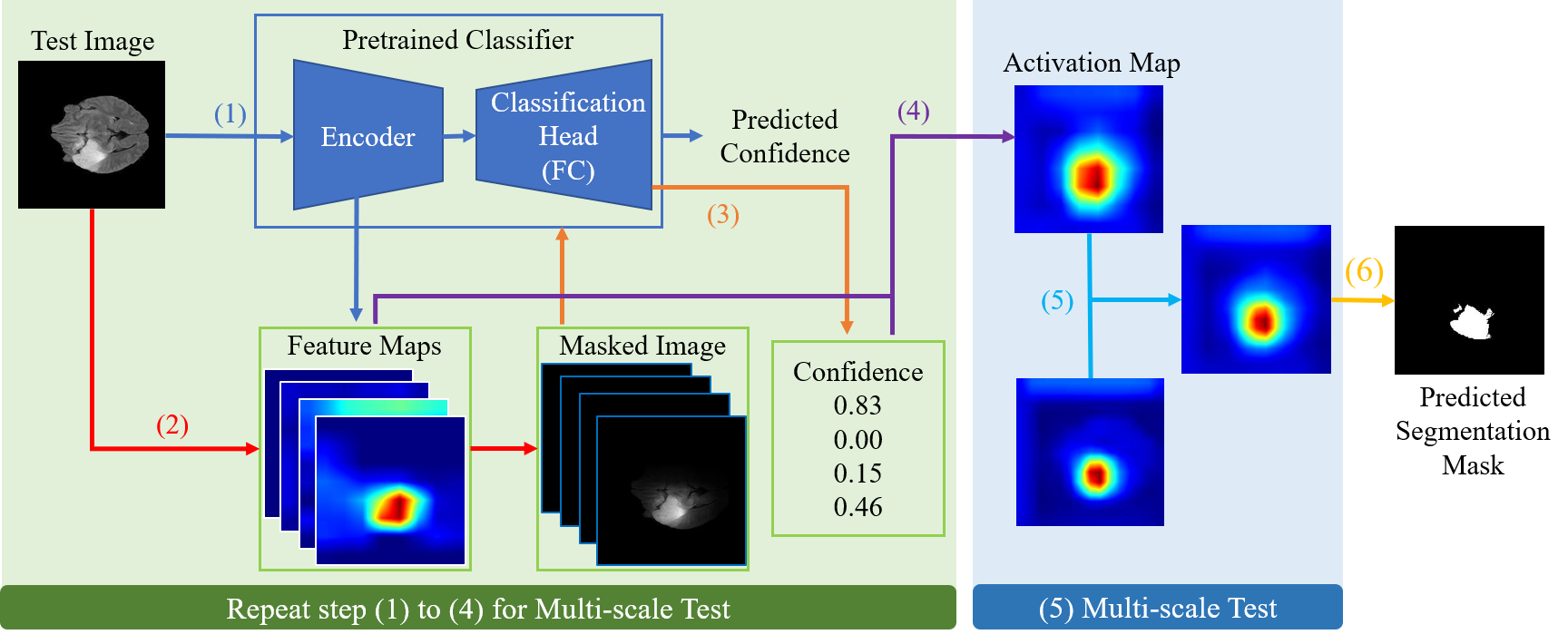}
\caption{Illustration of the proposed Cfd-CAM method. FC stands for the fully-connected layer.}
\label{Fig.Cfd-CAM}
\end{figure}

The Cfd-CAM method (Fig.~\ref{Fig.Cfd-CAM}) involves a binary classification network, e.g. ResNet-18 
\cite{he2016deep}, trained to determine whether an image contains a tumor (positive) or not (negative). For every test image, the trained classifier generates the corresponding segmentation mask using the proposed Cfd-CAM.

Similar to most CAM methods, the aggregated activation map is a linear combination of all the feature maps with a ReLU \cite{agarap2018deep} operation that clips all negative values to 0. However, to achieve accurate weights for feature maps, our proposed Cfd-CAM method constructs a confidence-induced algorithm to fit the contribution given by each feature map, which is introduced step-by-step as follows:

(1) We load the pretrained weight of the classifier model and obtain the feature maps of the input image from the last layer before global average pooling.

(2) For each feature map, we upsample it to the original image size and apply min-max scaling to normalize the value to [0,1]. Afterward, we mask every processed feature map to the original image.

(3) All the masked images are fed into the classification model to obtain its confidence. The importance of each feature map is determined as follows: if the confidence of the masked image is lower than the confidence of a blank image, then the result of the classification network, including the feature maps, shall not be trusted. In this case, we simply set the weight to $0$. Otherwise, the weight of the feature map is set as the confidence of the masked image.
To preserve the value of the generated map ranged [0,1], we apply the proportion to the weight for feature map aggregation.

(4) With the weight of each feature map determined, the aggregated activation map is generated.

(5) The proposed Cfd-CAM uses the activation maps obtained from the original image and the double-sized image for the experiment. To combine the knowledge given from different resolutions, the final activation map is calculated by the average of two activation maps, the original-sized and the double-sized (rescaled back to the original size).

(6) To generate the segmentation mask, we apply the Dense Conditional Random Field (DenseCRF) algorithm \cite{krahenbuhl2011efficient} to determine the pixel-wise prediction from the activation map.

\begin{figure}
\begin{center}
\includegraphics[width=0.9\linewidth]{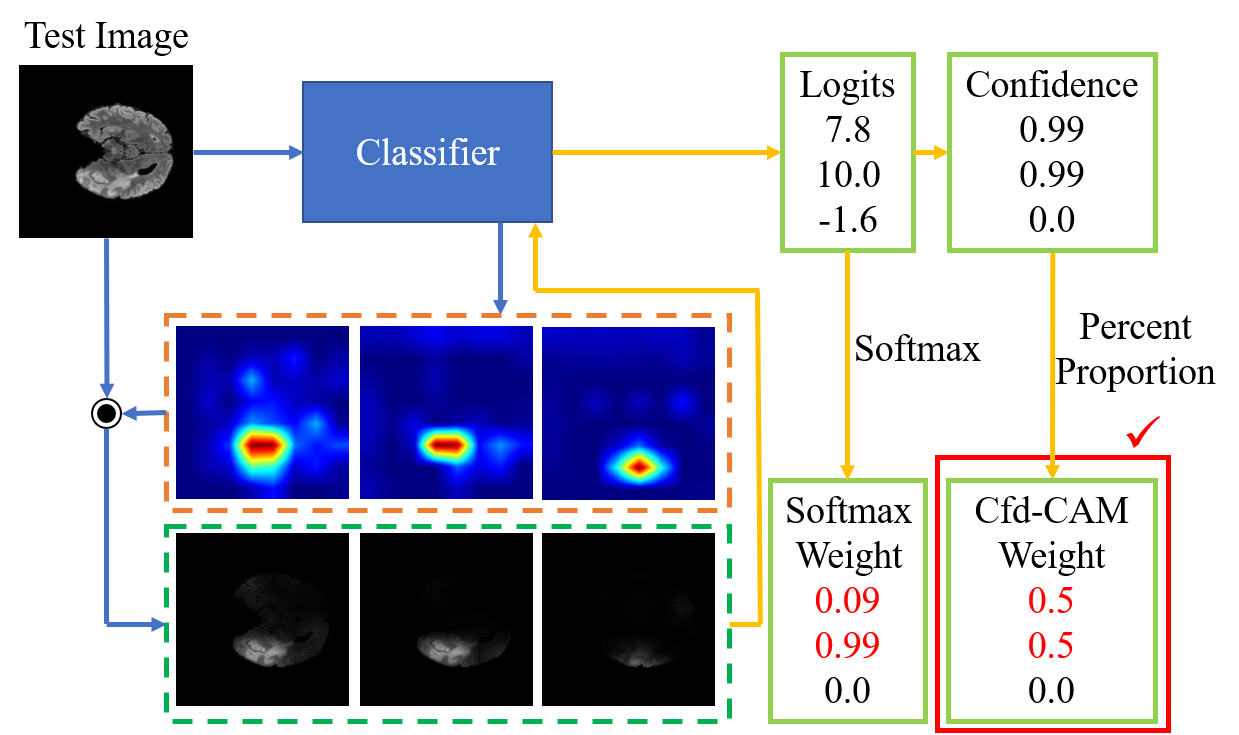}

\caption{Illustration of the weight modification of the previous work (softmax) and the proposed Cfd-CAM. The weights highlighted in red show the main difference of our Cfd-CAM.
}
\label{Fig.example}
\end{center}
\end{figure}

To demonstrate the effectiveness of our proposed Cfd-CAM, we provide an example to illustrate how it improves upon the previous method, as depicted in Fig.~\ref{Fig.example}. We consider a binary classification network that has three feature maps extracted before global average pooling. Assume the logits of the input masked by each feature map are ${7.8, 10.0, -1.6}$, respectively. If we were to use softmax or sigmoid activation with percentage proportions for the weight of each map, the results would be ${0.09, 0.90, 0.00}$ and ${0.5, 0.5, 0.00}$, respectively.

In the former method, the second feature map dominates the final activation map. However, the latter method suggests that the classification network should have similar confidence for the input masked by the first feature map and that by the second, implying that their contribution to the activation map should be equivalent.

It is more logical to use the network's confidence directly for the input masked by each feature map as the weight of that feature map. This approach is the basis of our proposed Cfd-CAM and shows its superiority over the previous method.

%% file: Section/Experiments.tex
\subsection{Dataset}

We evaluate our experiments on two MRI lesion segmentation datasets, the Brain Tumor Segmentation challenge (BraTS) \cite{menze2014multimodal,bakas2017advancing,bakas2018identifying} and the Cancer Genome Atlas Lower-Grade Glioma collection (TCGA-LGG) \cite{pedano2016radiology}.

The Brain Tumor Segmentation challenge (BraTS) dataset \cite{menze2014multimodal,bakas2017advancing,bakas2018identifying} contains 2,000 cases, each of which includes four 3D volumes from four different MRI modalities: T1, post-contrast enhanced T1 (T1-CE), T2, and T2 Fluid Attenuated Inversion Recovery (T2-FLAIR), as well as a corresponding segmentation ground-truth mask. The official data split divides these cases by the ratio of 8:1:1 for training, validation, and testing (5,802 positive and 1,073 negative images). In order to evaluate the performance, we use the validation set as our test set and report statistics on it. We preprocess the data by slicing each volume along the z-axis to form a total of 193,905 2D images, following the approach of Kang et al. \cite{kang2021towards} and Dey and Hong \cite{dey2021asc}. We use the ground-truth segmentation masks only in the final evaluation, not in the training process.

The TCGA-LGG dataset \cite{pedano2016radiology} consists of 110 patients with 3,600 2D images collected from the TCGA lower-grade glioma collection who had preoperative imaging data available, containing at least a fluid-attenuated inversion recovery (FLAIR) sequence. We follow the preprocess step and the split rate in the BraTS dataset to separate the dataset into training, validation, and testing (180 positive and 351 negative images) for experiments.

\subsection{Implementation Details and Evaluation Protocol}

We implement our method in PyTorch using a ResNet-18 model architecture for the classifier. We pretrain the classifier using SupCon \cite{khosla2020supervised} and fine-tune it in our experiments. Both pretraining and fine-tuning use the entire training set. The initial learning rate of the classifier is set to 1e-4 and follows the cosine annealing scheduler to decrease until the minimum learning rate is 5e-6, and we use the Adam optimizer. For model regularization, we set the weight decay to 1e-5. To maintain a fair comparison, all classifiers are trained until convergence with a test accuracy of over 0.9 for all image modalities. Note that only class labels are available in the training set.

We use the Dice score and Intersection over Union (IoU) to evaluate the quality of the semantic segmentation, following the approach of Xu et al. \cite{xu2019whole}, Tang et al. \cite{tang2021m}, and Qian et al. \cite{qian2022transformer}. In addition, we report the 95\% Hausdorff Distance (HD95) to evaluate the boundary of the prediction mask.

%% file: Section/Results.tex
\input{Table/Compare_CAM}

\subsection{Quantitative and Qualitative Comparison with State-of-the-art}
\label{result:cam_exp}

\begin{figure}
\begin{center}
\includegraphics[width=0.95\linewidth]{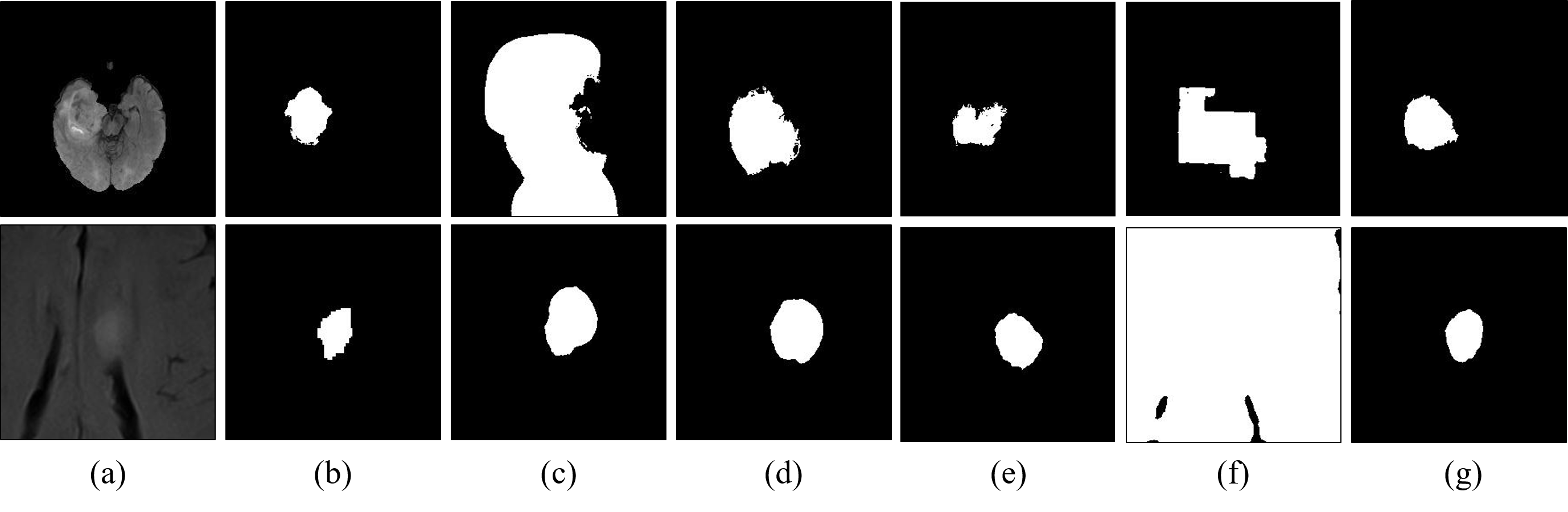}

\caption{Qualitative results of all methods using T2-FLAIR of BraTS dataset (first row) and the TCGA-LGG dataset (second row). (a) Input Image. (b) Ground Truth. (c) Grad-CAM \cite{selvaraju2017grad} (d) ScoreCAM \cite{wang2020score}. (e) LayerCAM \cite{jiang2021layercam}. (f) Swin-MIL \cite{qian2022transformer}. (g) Cfd-CAM (ours). 
}
\label{Fig.Result}
\vspace{-20pt}
\end{center}
\end{figure}

In this section, we evaluate the segmentation performance of the proposed Cfd-CAM and compare it with four state-of-the-art weakly-supervised segmentation methods: Grad-CAM \cite{selvaraju2017grad}, ScoreCAM \cite{wang2020score}, LayerCAM \cite{jiang2021layercam}, and Swin-MIL \cite{qian2022transformer}. To provide a comprehensive comparison, we also include an unsupervised approach C\&F \cite{chen2020medical}, its supervised version, and the supervised Optimized U-net \cite{futrega2021optimized}, which are non-CAM-based methods. Although the results from fully supervised and unsupervised methods are not directly comparable to weakly supervised CAM methods, we include them as interesting references for the potential performance ceiling and floor of all the CAM methods.

Table \ref{Table:Compare_Exp_CAM} presents the quantitative comparison results. Grad-CAM and ScoreCAM tend to overestimate the positive region and lead to false-positive segmentation, resulting in low dice and precision scores in the BraTS dataset. LayerCAM significantly improves the dice score to over 0.5 in the BraTS dataset by capturing activation maps at different resolutions. Swin-MIL also results in a dice score around 0.45, except for T2-FLAIR MRI in the BraTS dataset. Compared with other weakly supervised segmentation methods, the proposed Cfd-CAM achieves the highest dice score in all modalities of the BraTS dataset and the TCGA-LGG dataset.

Compared to the unsupervised baseline (UL), C\&F is unable to separate the tumor and the surrounding tissue due to low contrast, resulting in low dice scores in all experiments. With pixel-wise labels, the dice score of supervised C\&F improves significantly. Without any pixel-wise label, the proposed Cfd-CAM outperforms supervised C\&F in all modalities of the BraTS dataset.

The fully supervised (FSL) Optimized U-net achieves the highest dice score and IoU score in all experiments. However, even under different levels of supervision, there is still a performance gap between the weakly supervised CAM methods and the fully supervised state-of-the-art. This indicates that there is still potential room for WSSS methods to improve in the future.

Figure \ref{Fig.Result} shows the visualization of the segmentation results from all five weakly supervised approaches under two different datasets. Grad-CAM has the worst performance with a large falsely segmented region. Most of the boundaries are unacceptable due to the inappropriate construction of weights by back-propagation. Cfd-CAM significantly reduces the overestimation of the tumor area, resulting in improved weight construction through confidence. For the TCGA-LGG dataset, all six methods cover the exact tumor area, except Swin-MIL. The proposed Cfd-CAM results in the least area and fewer falsely segmented regions.

\input{Table/Ablation_Study}

\subsection{Ablation Study for Multi-scale Test}
\label{result:multi-scale}

\begin{figure}
\begin{center}
\includegraphics[width=0.9\linewidth]{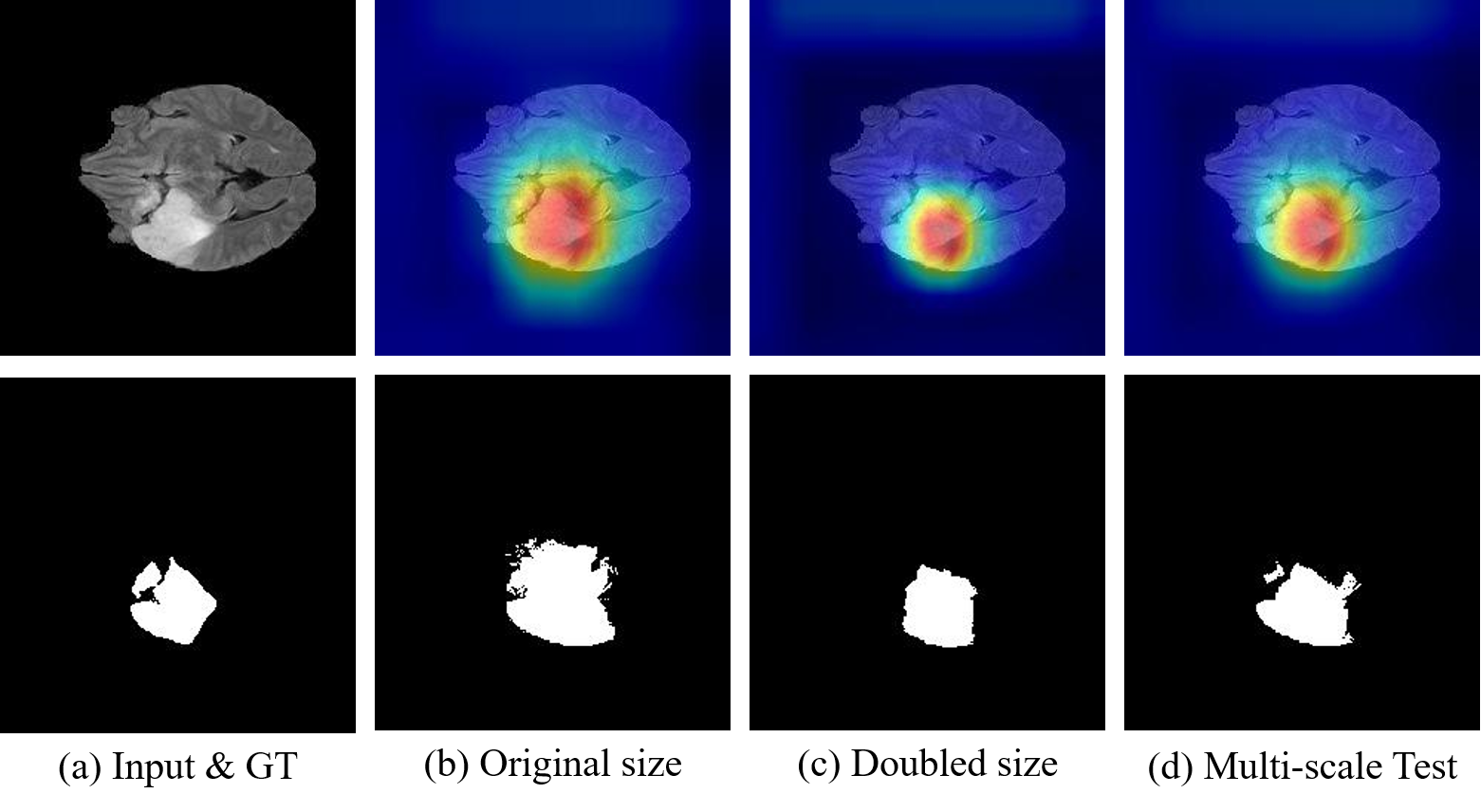}

\caption{Visualization of the activation map (top row) and ground-truth segmentation mask (bottom row) of Cfd-CAM from (b) single-scale test of the original size, (c) single-scale test of the doubled original size, and (d) multi-scale test. (a) represents the input image and its ground-truth (GT) segmentation mask. 
}
\label{Fig.Ablation}
\vspace{-20pt}
\end{center}
\end{figure}

In this section, we compare three different tests: single-scale test using original size, single-scale test using double the original size, and multi-scale test that combines advantages from different scales. Fig. \ref{Fig.Ablation} shows that the single-scale test using original size focuses on global vision and produces large activation regions, while the single-scale test with double the original size captures local and accurate boundaries, resulting in smaller activation regions. The multi-scale test, as shown in Fig. \ref{Fig.Ablation}(d), combines the advantages provided by different scales and produces segmentation closest to the ground truth.

Table \ref{Table:Ablation} presents the quantitative comparison results for BraTS and TCGA-LGG datasets, with the multi-scale test improving the segmentation dice score by about 8\% on average compared to the single-scale test using the original size. Overall, the multi-scale test outperforms the single-scale tests, providing more accurate segmentation.

%% file: Table/Compare_CAM.tex
\begin{table}[]
\centering
\caption{Comparison with weakly-unsupervised methods (WSSS), unsupervised method (UL), and fully supervised methods (FSL) on BraTS dataset with T1, T1-CE, T2, and T2-FLAIR MRI images and the TCGA-LGG dataset. Results are reported in the form of mean$\pm$std. We mark the highest score among WSSS methods with bold text.}
\label{Table:Compare_Exp_CAM}
\begin{tabular}{c|c|c|c|c|l}
\hline
Dataset & Type                  & Method            & Dice $\uparrow$ & IoU $\uparrow$  & \multicolumn{1}{c}{HD95 $\downarrow$} \\ \hline
\multirow{8}{*}{\shortstack{BraTS\\T1}} & \multirow{5}{*}{WSSS} & Grad-CAM (2016)   & 0.107$\pm$0.090    & 0.059$\pm$0.055    & 121.816$\pm$22.963                       \\
&                      & ScoreCAM (2020)   & 0.296$\pm$0.128    & 0.181$\pm$0.089    & 60.302$\pm$14.110                        \\
&                      & LayerCAM (2021)   & 0.571$\pm$0.170    & 0.419$\pm$0.161    & \textbf{23.335$\pm$27.369}                        \\
&                      & Swin-MIL (2022)   & 0.477$\pm$0.170    & 0.330$\pm$0.147    & 46.468$\pm$30.408                        \\
&                      & Cfd-CAM (ours)    & \textbf{0.575$\pm$0.169}    & \textbf{0.427$\pm$0.143}    & 23.859$\pm$20.243                        \\ \cline{2-6}
& UL                    & C\&F (2020)       & 0.200$\pm$0.082 & 0.113$\pm$0.051 & 79.187$\pm$14.304                     \\ \cline{2-6}
& \multirow{2}{*}{FSL}  & C\&F (2020)       & 0.572$\pm$0.196 & 0.426$\pm$0.187 & 29.027$\pm$20.881                     \\
&                      & Opt. U-net (2021) & 0.836$\pm$0.062 & 0.723$\pm$0.090 & 11.730$\pm$10.345                     \\ \hline
\multirow{8}{*}{\shortstack{BraTS\\T1-CE}} & \multirow{5}{*}{WSSS} & Grad-CAM (2016)   & 0.127$\pm$0.088    & 0.071$\pm$0.054    & 129.890$\pm$27.854                       \\
&                      & ScoreCAM (2020)   & 0.397$\pm$0.189    & 0.267$\pm$0.163    & 46.834$\pm$22.093                        \\
&                      & LayerCAM (2021)   & 0.510$\pm$0.209    & 0.367$\pm$0.180    & 29.850$\pm$45.877                        \\
&                      & Swin-MIL (2022)   & 0.460$\pm$0.169    & 0.314$\pm$0.140    & 46.996$\pm$22.821                        \\
&                      & Cfd-CAM (ours)    & \textbf{0.610$\pm$0.156}    & \textbf{0.456$\pm$0.157}    & \textbf{23.352$\pm$13.642}                        \\ \cline{2-6}
& UL                    & C\&F (2020)       & 0.179$\pm$0.080 & 0.101$\pm$0.050 & 77.982$\pm$14.042                     \\ \cline{2-6}
& \multirow{2}{*}{FSL}  & C\&F (2020)       & 0.246$\pm$0.104 & 0.144$\pm$0.070 & 130.616$\pm$9.879                     \\
&                      & Opt. U-net (2021) & 0.845$\pm$0.058 & 0.736$\pm$0.085 & 11.593$\pm$11.120                     \\ \hline
\multirow{8}{*}{\shortstack{BraTS\\T2}} & \multirow{5}{*}{WSSS} & Grad-CAM (2016)   & 0.049$\pm$0.058    & 0.026$\pm$0.034    & 141.025$\pm$23.107                       \\
&                      & ScoreCAM (2020)   & 0.530$\pm$0.184    & 0.382$\pm$0.174    & 28.611$\pm$11.596                        \\
&                      & LayerCAM (2021)   & 0.624$\pm$0.178    & 0.476$\pm$0.173    & 23.978$\pm$44.323                        \\
&                      & Swin-MIL (2022)   & 0.437$\pm$0.149    & 0.290$\pm$0.117    & 38.006$\pm$30.000                        \\
&                      & Cfd-CAM (ours)    & \textbf{0.660$\pm$0.132}    & \textbf{0.506$\pm$0.141}    & \textbf{19.323$\pm$10.714}                        \\ \cline{2-6}
& UL                    & C\&F (2020)       & 0.230$\pm$0.089 & 0.133$\pm$0.058 & 76.256$\pm$13.192                     \\ \cline{2-6}
& \multirow{2}{*}{FSL}  & C\&F (2020)       & 0.611$\pm$0.221 & 0.474$\pm$0.217 & 109.817$\pm$27.735                    \\
&                      & Opt. U-net (2021) & 0.884$\pm$0.064 & 0.798$\pm$0.098 & 8.349$\pm$9.125                       \\ \hline
\multirow{8}{*}{\shortstack{BraTS\\T2-FLAIR}} & \multirow{5}{*}{WSSS} & Grad-CAM (2016)   & 0.150$\pm$0.077    & 0.083$\pm$0.050    & 110.031$\pm$23.307                       \\
&                      & ScoreCAM (2020)   & 0.432$\pm$0.209    & 0.299$\pm$0.178    & 39.385$\pm$17.182                        \\
&                      & LayerCAM (2021)   & 0.652$\pm$0.206    & 0.515$\pm$0.210    & 22.055$\pm$33.959                        \\
&                      & Swin-MIL (2022)   & 0.272$\pm$0.115    & 0.163$\pm$0.079    & 41.870$\pm$19.231                        \\
&                      & Cfd-CAM (ours)    & \textbf{0.733$\pm$0.126}    & \textbf{0.594$\pm$0.153}    & \textbf{15.273$\pm$8.251}                         \\ \cline{2-6}
& UL                    & C\&F (2020)       & 0.306$\pm$0.190 & 0.199$\pm$0.167 & 75.651$\pm$14.214                     \\ \cline{2-6}
& \multirow{2}{*}{FSL}  & C\&F (2020)       & 0.578$\pm$0.137 & 0.419$\pm$0.130 & 138.138$\pm$14.283                    \\
&                      & Opt. U-net (2021) & 0.914$\pm$0.058 & 0.847$\pm$0.093 & 8.093$\pm$11.879                      \\ \hline
\multirow{8}{*}{TCGA-LGG} & \multirow{5}{*}{WSSS} & Grad-CAM (2016)   & 0.407$\pm$0.229          & 0.283$\pm$0.190          & 81.278$\pm$38.648                        \\
&                      & ScoreCAM (2020)   & 0.490$\pm$0.222          & 0.351$\pm$0.184          & 57.821$\pm$30.275                        \\
&                      & LayerCAM (2021)   & 0.378$\pm$0.211          & 0.255$\pm$0.171          & 69.696$\pm$42.493                        \\
&                      & Swin-MIL (2022)   & 0.271$\pm$0.144          & 0.165$\pm$0.101          & 111.634$\pm$17.670                       \\
&                      & Cfd-CAM (ours)    & \textbf{0.508$\pm$0.223} & \textbf{0.369$\pm$0.193} & \textbf{56.866$\pm$29.337}               \\ \cline{2-6}
& UL                    & C\&F (2020)       & 0.300$\pm$0.162          & 0.188$\pm$0.119          & 101.980$\pm$18.358                       \\ \cline{2-6}
& \multirow{2}{*}{FSL}  & C\&F (2020)       & 0.637$\pm$0.216          & 0.502$\pm$0.224          & 46.925$\pm$17.889                        \\
&                      & Opt. U-Net (2021) & 0.668$\pm$0.196          & 0.532$\pm$0.209          & 47.035$\pm$19.529                        \\ \hline
\end{tabular}
\end{table}

%% file: Table/Ablation_Study.tex
\begin{table}
\centering
\caption{Ablation study of single- (ss) and multi-scale (ms) test of the proposed Cfd-CAM using both BraTS dataset and TCGA-LGG dataset. Results are reported in the form of mean$\pm$std.}
\label{Table:Ablation}
\begin{tabular}{l|l|c|c|l}
\hline
Dataset                         & \multicolumn{1}{c|}{Method} & Dice $\uparrow$       & IoU $\uparrow$        & \multicolumn{1}{c}{HD95 $\downarrow$} \\ \hline
\multirow{2}{*}{BraTS T1}       & Cfd-CAM (ss)                & 0.406$\pm$0.152          & 0.266$\pm$0.121          & 45.610$\pm$15.356                        \\
                                & Cfd-CAM (ms)                & \textbf{0.575$\pm$0.169} & \textbf{0.427$\pm$0.143} & \textbf{23.859$\pm$20.243}               \\ \hline
\multirow{2}{*}{BraTS T1-CE}    & Cfd-CAM (ss)                & 0.537$\pm$0.189          & 0.389$\pm$0.176          & 30.075$\pm$18.408                        \\
                                & Cfd-CAM (ms)                & \textbf{0.610$\pm$0.156} & \textbf{0.456$\pm$0.157} & \textbf{23.352$\pm$13.642}               \\ \hline
\multirow{2}{*}{BraTS T2}       & Cfd-CAM (ss)                & 0.573$\pm$0.179          & 0.424$\pm$0.177          & 24.716$\pm$9.795                         \\
                                & Cfd-CAM (ms)                & \textbf{0.646$\pm$0.142} & \textbf{0.493$\pm$0.154} & \textbf{19.696$\pm$7.494}                \\ \hline
\multirow{2}{*}{BraTS T2-FLAIR} & Cfd-CAM (ss)                & 0.599$\pm$0.221          & 0.462$\pm$0.221          & 24.604$\pm$14.609                        \\
                                & Cfd-CAM (ms)                & \textbf{0.698$\pm$0.159} & \textbf{0.557$\pm$0.181} & \textbf{17.675$\pm$9.698}                \\ \hline
\multirow{2}{*}{TCGA-LGG}       & Cfd-CAM (ss)                & 0.503$\pm$0.253          & 0.373$\pm$0.226          & 58.668$\pm$35.295                        \\
                                & Cfd-CAM (ms)                & \textbf{0.508$\pm$0.223} & \textbf{0.369$\pm$0.193} & \textbf{56.866$\pm$29.337}               \\ \hline
\end{tabular}
\end{table}

%% file: Section/Conclusion.tex
We introduce Cfd-CAM, a novel brain tumor segmentation method for MRI images that utilizes only class labels. By weighting feature maps with confidence, our approach generates an activation map that overcomes limitations of other CAM-based methods. Experimental results on the BraTS and TCGA-LGG datasets show that Cfd-CAM outperforms other weakly-supervised segmentation approaches based on CAM, indicating its effectiveness in accurately segmenting brain tumors based solely on class labels.

%% file: Table/Supple_Res50.tex
\begin{table}
\centering
\caption{Comparison with WSSS Methods using ResNet-50 on BraTS dataset with T1, T1-CE, T2, and T2-FLAIR MRI images. Results are reported in the form of mean$\pm$std. We mark the highest score among WSSS methods with bold text.}
\label{Table:Supple_Res50}
\begin{tabular}{c|c|c|l}
\hline
\multicolumn{4}{c}{BraTS T1} \\ \hline
Method          & Dice $\uparrow$       & IoU $\uparrow$        & \multicolumn{1}{c}{HD95 $\downarrow$} \\ \hline
Grad-CAM (2016) & 0.046$\pm$0.086          & 0.026$\pm$0.053          & 132.469$\pm$33.056                       \\
ScoreCAM (2020) & 0.233$\pm$0.089          & 0.134$\pm$0.058          & 75.245$\pm$15.082                        \\
LayerCAM (2021) & 0.453$\pm$0.195          & 0.313$\pm$0.164          & 40.190$\pm$55.079                        \\
Cfd-CAM  (ours) & \textbf{0.479$\pm$0.173} & \textbf{0.332$\pm$0.149} & \textbf{35.675$\pm$26.325}               \\ \hline
\multicolumn{4}{c}{BraTS T1-CE} \\ \hline
Method          & Dice $\uparrow$       & IoU $\uparrow$        & \multicolumn{1}{c}{HD95 $\downarrow$} \\ \hline
Grad-CAM (2016) & 0.341$\pm$0.132          & 0.213$\pm$0.096          & 48.055$\pm$17.094                        \\
ScoreCAM (2020) & 0.343$\pm$0.235          & 0.234$\pm$0.188          & 50.868$\pm$32.598                        \\
LayerCAM (2021) & 0.480$\pm$0.213          & \textbf{0.341$\pm$0.176} & 40.003$\pm$72.395                        \\
Cfd-CAM  (ours) & \textbf{0.487$\pm$0.176} & 0.340$\pm$0.156          & \textbf{34.544$\pm$21.150}               \\ \hline
\multicolumn{4}{c}{BraTS T2} \\ \hline
Method          & Dice $\uparrow$       & IoU $\uparrow$        & \multicolumn{1}{c}{HD95 $\downarrow$} \\ \hline
Grad-CAM (2016) & 0.075$\pm$0.060          & 0.040$\pm$0.035          & 147.272$\pm$26.227                       \\
ScoreCAM (2020) & 0.479$\pm$0.219          & 0.342$\pm$0.193          & 38.509$\pm$27.572                        \\
LayerCAM (2021) & 0.638$\pm$0.184          & 0.493$\pm$0.190          & 26.907$\pm$33.535                        \\
Cfd-CAM  (ours) & \textbf{0.694$\pm$0.162} & \textbf{0.554$\pm$0.185} & \textbf{17.659$\pm$8.104}                \\ \hline
\multicolumn{4}{c}{BraTS T2-FLAIR} \\ \hline
Method          & Dice $\uparrow$       & IoU $\uparrow$        & \multicolumn{1}{c}{HD95 $\downarrow$} \\ \hline
Grad-CAM (2016) & 0.221$\pm$0.084          & 0.127$\pm$0.056          & 68.666$\pm$22.125                        \\
ScoreCAM (2020) & 0.588$\pm$0.252          & 0.458$\pm$0.239          & 26.126$\pm$23.682                        \\
LayerCAM (2021) & 0.638$\pm$0.196          & 0.497$\pm$0.200          & 23.290$\pm$35.905                        \\
Cfd-CAM  (ours) & \textbf{0.746$\pm$0.131} & \textbf{0.611$\pm$0.161} & \textbf{14.967$\pm$8.135}                \\ \hline
\end{tabular}
\end{table}

%% file: Table/Supple_cfd.tex
\begin{table}
\centering
\caption{Ablation study for the proposed method using logits and confidence-weighted for feature map aggregation on BraTS dataset with T1, T1-CE, T2, and T2-FLAIR MRI images. Results are reported in the form of mean$\pm$std.}
\label{Table:Ablation}
\begin{tabular}{l|l|c|c|l}
\hline
Dataset                         & Method     & Dice $\uparrow$       & IoU $\uparrow$        & \multicolumn{1}{c}{HD95 $\downarrow$} \\ \hline
\multirow{2}{*}{BraTS T1}       & Logits     & 0.171$\pm$0.196          & 0.108$\pm$0.135          & 164.052$\pm$147.436                      \\
                                & Confidence & \textbf{0.575$\pm$0.169} & \textbf{0.427$\pm$0.143} & \textbf{23.859$\pm$20.243}               \\ \hline
\multirow{2}{*}{BraTS T1-CE}    & Logits     & 0.528$\pm$0.191          & 0.381$\pm$0.170          & 32.063$\pm$30.751                        \\
                                & Confidence & \textbf{0.610$\pm$0.156} & \textbf{0.456$\pm$0.157} & \textbf{23.352$\pm$13.642}               \\ \hline
\multirow{2}{*}{BraTS T2}       & Logits     & 0.639$\pm$0.152          & 0.488$\pm$0.161          & 21.301$\pm$24.374                        \\
                                & Confidence & \textbf{0.660$\pm$0.132} & \textbf{0.506$\pm$0.141} & \textbf{19.323$\pm$10.714}               \\ \hline
\multirow{2}{*}{BraTS T2-FLAIR} & Logits     & 0.639$\pm$0.211          & 0.501$\pm$0.208          & 32.332$\pm$66.499                        \\
                                & Confidence & \textbf{0.733$\pm$0.126} & \textbf{0.594$\pm$0.153} & \textbf{15.273$\pm$8.251}                \\ \hline
\end{tabular}
\end{table}